\documentclass[acmsmall,screen,review=false]{acmart}
\AtBeginDocument{%
  }

\acmJournal{TAP}
\acmVolume{0}
\acmNumber{0}
\acmArticle{0}
\acmYear{2026}
\acmMonth{1}
\acmDOI{10.1145/nnnnnnn.nnnnnnn}

\copyrightyear{2026}
\setcopyright{acmlicensed}

\usepackage{booktabs}

\usepackage{xcolor}
\newif\ifTODOS \TODOStrue

\begin{document}

\title{Effects of Conventional, Adaptive, and AI-Powered Upscaling on the Legibility of License Plates}

\author{Hany Farid}
\affiliation{%
  \institution{Dartmouth College}
  \city{Hanover}
  \state{NH}
  \country{USA}
}
\email{hany.farid@dartmouth.edu}

\author{Joshua D. Bucci}
\affiliation{%
  \institution{Dartmouth College}
  \city{Hanover}
  \state{NH}
  \country{USA}
}
\email{joshua.d.bucci@dartmouth.edu}

\author{Emily A. Cooper}
\correspondingauthor
\affiliation{%
  \institution{Dartmouth College}
  \city{Hanover}
  \state{NH}
  \country{USA}
}
\email{emily.a.cooper@dartmouth.edu}


\begin{abstract}
Small details in digital images --- a face, a street sign, a license plate --- are routinely upscaled in hopes of making them interpretable. Upscaling options now range from conventional interpolation, whose operations are spatially invariant and fully explainable, to AI-powered methods that synthesize detail potentially absent from the source. Here, we ask which of these techniques most reliably helps human readers. In two online experiments, we measured legibility of $255$ US license plates ranging in resolution from $128$ to $8$ pixels tall, as well as their legibility after upscaling by eight different methods. Legibility of the un-upscaled plates collapsed from ceiling to floor across a narrow band of plate heights ($32$ to $10$ pixels). Conventional upscaling techniques (bilinear, bicubic, and Lanczos interpolation) gave modest gains of about $5$-$9$ percentage points within this range. Nearest-neighbor interpolation, however, was harmful, rendering unreadable many plates that were readable without any upscaling. Advanced methods diverged: SwinIR, an open-source AI enhancer, raised performance by up to $38$ points relative to no upscaling, while an adaptive denoiser-based image prior and commercial AI upscaler tended to impair legibility. We found no strong evidence of consistent character hallucinations across methods. Decomposing the task, conventional interpolators mainly helped readers count characters, whereas SwinIR alone made the characters themselves more identifiable. 
\end{abstract}

\ccsdesc[500]{Applied computing~Psychology}
\ccsdesc[500]{Applied computing~Law}
\ccsdesc[500]{Computing methodologies~Image processing}

\keywords{image processing, upscaling, super-resolution, legibility, digital forensics, human performance}


\begin{teaserfigure}
 \includegraphics[width=\textwidth]{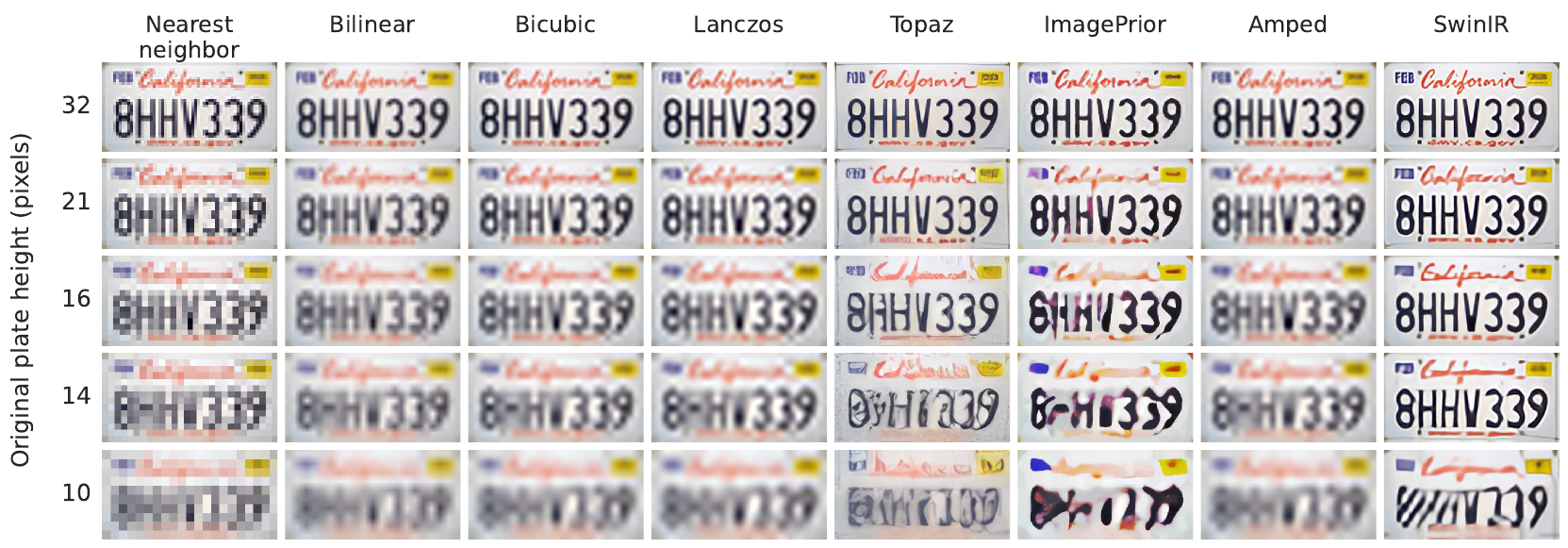}
    \caption{A single license plate at five original heights, from $32$ pixels down to $10$, each upscaled back to high resolution with a range of conventional, adaptive, and AI-powered upscaling techniques. Reading down a column shows how legibility is lost as the plate shrinks; across a row shows how differently the methods treat the same input.}
    \label{fig:teaser}
\end{teaserfigure}

\maketitle

\section{Introduction}

Digital images are rarely presented to a human viewer exactly as they were recorded. Between capture and display sit enhancement operations that adjust contrast, suppress noise, sharpen edges, and change size. Sometimes these enhancements operate purely for aesthetics. However, in legal, journalistic, and other investigative contexts these enhancements can operate in the service of making the content easier for a person to interpret veridically. 

Conventional enhancement techniques  are fixed functions of the input pixels. These techniques include interpolation kernels, filters, and tone adjustments, long available in image editors. They redistribute and rescale what is already there, producing an image that is smoother or of higher contrast. Newer enhancement methods, now trained on large collections of photographs and sometimes powered by artificial intelligence (AI), can go beyond conventional techniques to synthesize detail that is plausible for natural images~\cite{ledig2017photo,menon2020pulse,kadkhodaie2021stochastic}.

Which enhancement technique is best at supporting improved human interpretation is ultimately a question about a human observer. Recent studies suggest that, while AI-powered image enhancements may appear subjectively superior, they do not necessarily improve human performance on image interpretation. For example, a recent study compared people's face recognition performance from small images that were upscaled with either conventional bicubic interpolation or two AI-powered super resolution techniques (SRFlow and GLEAN)~\cite{velan2022does,lugmayr2020srflow,chan2021glean}. Neither AI-powered technique resulted in better recognition performance than bicubic interpolation. In another example, however, radiologists were asked to localize visual features indicative of breast cancer in a series of medical images, some of which were enhanced with an AI-powered denoiser~\cite{chan2023deep}. The denoised images supported improved detection performance relative to standard images. The improvement, however, was not benchmarked against a conventional, explainable denoiser. Given the vast space of images and human image interpretation tasks, the limited work in this space does not yet support generalizable conclusions about whether AI-powered enhancements are superior to the conventional toolkit.

Here, we evaluate how various image upscaling techniques --- that is, techniques that increase the size of a small image by interpolating pixels --- affect the human legibility of text on license plates. We chose license plate reading because the stimulus set is well-defined and applicable to real-world situations, and the perceptual constraints of reading are well-characterized. For instance, letter identification relies on a narrow band of spatial frequencies (approximately 2–3 cycles per character), causing legibility to fail predictably when a character is under-sampled~\cite{solomon1994visual,sheedy2005text}. 

In a large-scale online study, we examined the effect of a range of image upscaling techniques on human readability of license plates. Example images from each of these techniques are shown in Figure~\ref{fig:teaser}. We used conventional interpolation kernels, a denoiser-based image prior, a learning-based super-resolution network, and two commercial products (one AI-powered and one purpose-built for forensic analysis).

\section{Methods}

\subsection*{License Plate Images}

\begin{figure}[t]
\centering
\includegraphics[width=\textwidth]{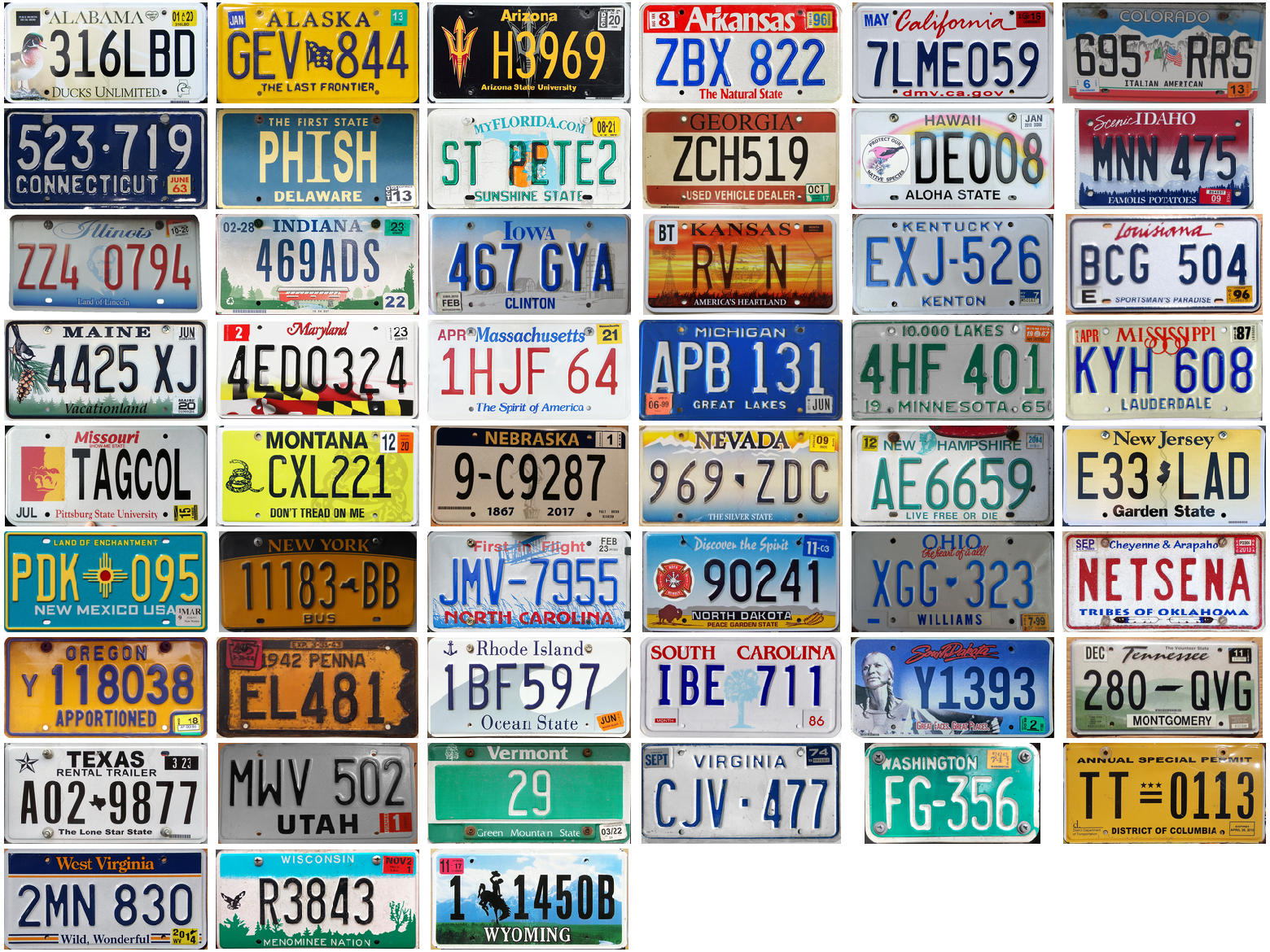}
\caption{Examples of one license plate for each locale used in the experiments. Source: \cite{dincer_us_license_plates}, CC BY-SA 4.0.}
\label{fig:plates}
\end{figure}

We first created a dataset of high-quality license plate images with known characters (Figure~\ref{fig:plates}). We started with a publicly-available US License Plates dataset~\cite{dincer_us_license_plates}. For each of the 50 states in the US and Washington, DC, we selected the first five images that (1) were modern enough to have a visible registration sticker, (2) were at least $512$ pixels tall, and (3) had no vertically stacked letters. New Jersey and New York plates in the dataset did not have registration stickers, so we selected the top five that met the other criteria. If this process yielded less than five for a given state, we augmented with images from~\cite{lorch2019forensic} or from images available online. This yielded $255$ unique plates. For each plate, we then manually specified a tight bounding box around the alpha-numeric plate characters and recorded the ground-truth characters. In addition, we denoted the presence of graphics, spaces, dashes, and dots.

All original plate images were resized to a height of $128$ pixels and saved as uncompressed PNG files. This served as the ``full resolution'' version of each plate, as there was more than sufficient resolution for clear legibility of all characters when viewed on a typical computer display at a typical viewing distance~\cite{sheedy2005text}. We then downscaled each image by factors of $2$, $3$, $4$, $6$, $8$, $9$, $12$, and $16$, yielding images that were $64$, $42$, $32$, $21$, $16$, $14$, $10$, and $8$ pixels tall (non-integer heights were rounded down). These heights refer to the plate image, not the characters. Scale factors were selected based on pilot data to span the range from fully legible to not legible at all. 

Each downscaled image was upscaled back to full resolution ($128$ pixels) using a range of image upscaling methods, described in the next section. In Experiment 1, we upscaled using conventional, spatially-invariant interpolation algorithms. In Experiment 2, we asked whether spatially-variant methods---adaptive, AI-powered, and commercial---affect legibility.

\subsection*{Stimuli: Experiment 1}

The four upscaling methods used in Experiment 1 all rely on fixed interpolation kernels. Each is fully deterministic and explainable, and each draws solely on the pixels of the image being upscaled:
\begin{itemize}
\item \textit{Nearest neighbor} assigns each output pixel the value of the single closest input pixel. 
\item \textit{Bilinear} interpolation computes each output pixel as a distance-weighted average of the four nearest input pixels. 
\item \textit{Bicubic} interpolation fits a cubic polynomial over the sixteen nearest input pixels. 
\item \textit{Lanczos} resampling convolves the image with a windowed sinc kernel.
\end{itemize}

Together these techniques span the practical range from no new pixel values (nearest neighbor) to near-optimal band-limited reconstruction (Lanczos). For each method, we used the implementation in the Python Imaging Library (PIL). For a baseline comparison, we also included the original full resolution images and ones that were not upscaled (i.e.,~the downsized images between $64$ and $8$ pixels tall).

\subsection*{Stimuli: Experiment 2}

Experiment 2 introduced new upscaling methods that are adaptive to the image content and can synthesize detail not recoverable from the source pixels alone: 
\begin{itemize}
\item \textit{Image Prior}~\cite{kadkhodaie2021stochastic} upscales images using a prior over natural images learned implicitly from a trained denoiser, returning a sample from the distribution of high-resolution images consistent with the low-resolution input. We histogram-matched the resulting images back to the original.
\item \textit{SwinIR}~\cite{liang2021swinir} is a transformer-based super-resolution network trained on natural image datasets, representative of the current deep-learning state of the art in the research literature. SwinIR was selected among several open-source learning-based candidates following an examination of the outputs using automated optical character recognition (OCR).
\item \textit{Topaz (v.1.2.2)} is a widely available commercial AI-upscaling product. We used this tool via the available plug-in for Adobe Photoshop.
\item \textit{Amped Smart Resize} is an upscaling method in Amped FIVE, a commercial image and video enhancement suite for forensic use~\cite{ramponi1999warped}. It draws only on the pixels of the image being upscaled, but unlike the Experiment 1 methods its interpolation is space-variant and data-dependent. It was included because it reflects an advanced option available in current forensic practice.
\end{itemize}
Bilinear interpolation and the baseline un-upscaled images from Experiment 1 were included again as anchors common to both experiments.

\subsection*{Participants}

Both experiments used the same procedure. Participants were recruited through Prolific and completed the study online in a web browser. Study participation was limited to US residents whose primary language was English with self-reported normal vision and an approval rating above $90\%$. Participants were required to complete the experiments on a laptop or desktop computer (no mobile devices). As non-native display resolutions could result in uncontrolled secondary pixel interpolation, participants were instructed only to select the study if they were willing and able to check and adjust their display resolution (see more details below). Each experiment comprised five survey versions of $125$ participants each ($n=625$ per experiment), giving $1{,}250$ submissions in total. All participants gave informed consent and were compensated for their time.

\subsection*{Procedure}

Before the plate reading task, participants reported their age, gender, whether they were using a laptop or a desktop computer, and their operating system. They were then instructed to verify that their display was set to its native resolution, given operating-system-specific instructions for doing so, and asked to report the resolution they were using. 

Each survey version included one plate (out of five) from each of the $51$ locales, resulting in $51$ trials. Across the five versions all $255$ plates were shown. The order of the $51$ plates was randomized for each participant. The upscaling method and downscaling factor assigned to each plate were randomized across participants. Participants typed the characters they could read into a free-text box; capitalization was stated to be unimportant, and participants were told that some images would be small or blurry and that they should do their best.

The majority of participants completed only one survey version, however twenty-six people in Experiment 1 and three people in Experiment 2 participated in two survey versions. Each survey contained distinct license plates, so recall from one version did not help performance on another.

\subsection*{Data Preparation and Exclusions}

Responses were normalized before scoring by removing all characters outside A--Z and 0--9. Responses indicating that the participant could not read the plate (for example ``cannot see'', ``too blurry'', ``n/a'', ``????''), together with any response that contained no alphanumeric characters, were coded as non-attempts and scored as incorrect. These were identified by matching against an explicit list of stock phrases, compiled by inspecting the responses. These responses accounted for $0.30\%$ of retained trials in Experiment 1 and $0.32\%$ in Experiment 2.

After this response screening, participants were excluded on three grounds. First, one participant in Experiment 2 was excluded for reporting that they had zoomed into the images, contrary to instructions. Second, participants whose reported display resolution could not be interpreted with both an identifiable width and height, or which indicated a mobile device, were excluded ($79$ in Experiment 1, $78$ in Experiment 2). Third, we computed each remaining participant's response accuracy (rate of responses that exactly matched ground truth, ignoring letter case) on the trials where the plate was least degraded (downscaling factor $\leq 3$, i.e. $\geq 42$ pixels tall) and excluded those falling below the $5^{th}$ percentile of that distribution. This threshold was computed within each experiment separately ($87.3\%$ in Experiment 1, $86.5\%$ in Experiment 2), resulting in the exclusion of $27$ participants in each Experiment. This screening left $519$ participants (out of the original $625$) and $26{,}469$ trials in each Experiment.

The retained participants had a mean age of $39.4$ years (standard deviation $11.5$) in Experiment 1 and $41.1$ years (standard deviation $12.9$) in Experiment 2 (one participant in each experiment did not have a reported age). Participants were approximately evenly split between male/female genders (both Experiments: $51\%$ male, $48\%$ female, $1\%$ non-binary or prefer not to say). 

Reported screen resolutions were dominated by $1920 \times 1080$ (approximately $50\%$ of retained participants). The full range spanned $1176 \times 664$ ($0.78$ megapixels) to $5120 \times 2160$ ($11.06$ megapixels). We observed no reliable differences in overall performance as a function of screen resolution, so we analyze all resolutions together. 

The median completion time was $13.7$ minutes in Experiment 1 and $14.3$ minutes in Experiment 2. 

\subsection*{Outcome Measures}

Our primary measure was the percentage of trials on which the participant response was character-for-character identical to the ground truth (i.e.,~an exact match). Because several character pairs are ambiguous even at good resolution, we also computed the percentage of almost-exact matches, in which a small set of commonly confusable characters were treated as interchangeable. 

The set of commonly confusable characters was determined from the data. Restricting to trials in the no-upscaling condition across both experiments, we tabulated character substitutions on incorrect responses whose length matched the ground truth. Pairs were ranked by confusion rate: for a pair of characters, the number of times either was read as the other, divided by the total number of errors made on either. Allowing for substitutions of the five highest-rate pairs ($E/F$, $5/S$, $0/D$, $0/O$, and $8/B$), we recomputed the percentage of correct trials. As $D$ and $O$ were both interchangeable with $0$, we also allowed $D/O$ confusions.

The almost-exact criterion raised match rates only slightly, by two to three percentage points overall ($58.9\%$ to $61.6\%$ in Experiment 1, $60.6\%$ to $62.6\%$ in Experiment 2). The main conclusions reported below are the same under either measure although some individual statistical comparisons differ. For simplicity, therefore, we report exact-match results only.

\subsection*{Main Analysis}

For the data visualizations, exact match percentages were computed at the level of individual trials. Within each combination of upscaling method and plate height, we report the percentage of all trials scored correct, with $95\%$ confidence intervals. Differences between each method and the no-upscaling baseline were computed as differences in percentages. 

These visualizations do not account for the clustering of trials within participants or within plates. For statistical inference, we fit one mixed-effects logistic regression per experiment. The model predicted the log odds of a correct response from upscaling method, original plate height (log transformed and centered on the mean across modeled heights), and their interaction, with random intercepts for participant and plate. No upscaling served as the reference method. Fitted coefficients were exponentiated to convert to odds ratios. Each method coefficient gives that method's odds ratio against no upscaling at the center of the height range. The height coefficient gives the change in odds per doubling of plate height without upscaling. Each interaction coefficient gives the factor by which that method's odds ratio changes per doubling.

Models were fitted by maximum likelihood with the Laplace approximation in MATLAB's \texttt{fitglme}. The analysis was restricted to the original plate heights at which performance was neither at ceiling nor at floor: $32$, $21$, $16$, $14$ and $10$ pixels. Above and below these heights, all methods resulted in performance above $96\%$ and below $1\%$, respectively. For all statistical tests, we use a significance threshold of $p \leq 0.05$.

In a follow-up analysis, we compared each method with no upscaling at each of the five modeled heights. Each comparison is an odds ratio computed directly from the trial counts. Because every plate appears about two to three times in each method/height combination and plates differ in difficulty, trials from the same plate are not independent. We therefore computed standard errors that are robust to clustering by plate~\cite{williams2000note}. Participants contributed fewer than two trials on average to each method/height combination, and the mixed-effects models attributed far less variance to participants than to plates, so participant identity was not included in the clustering. Using these standard errors, we report a $95\%$ confidence interval and two-sided z-test for each odds ratio. This approach yielded $20$ comparisons in Experiment 1 (four methods × five heights) and $25$ in Experiment 2 (five methods × five heights). We controlled the false discovery rate within each experiment with the Benjamini–Hochberg procedure and report the adjusted p-values as $p_c$. Note that odds ratios and percentage-point differences can rank comparisons differently. 

\subsection*{Exploratory analysis: Hallucination testing}

We asked whether any upscaling method was more likely to make readers ``hallucinate'' a character, that is, to read a specific wrong character reliably, as a synthesized detail might cause, rather than to guess at random. So that we could match individual responses to individual characters, in this analysis, we only examined responses whose length matched the ground truth. The unit of analysis was one character position on one plate as rendered by one method. Positions were kept distinct rather than combined by character class, because how a character appears after upscaling depends on the plate and on its neighbors. The $255$ plates thus contributed $1{,}494$ distinct character positions, spanning all $36$ alphanumeric characters. Each position was evaluated separately for each of the eight upscaling methods and for no upscaling, pooling responses over the modeled heights ($32$--$10$ pixels), yielding $13{,}446$ position-by-method units.

For each position we asked what the majority answer would be for a group of $5$ readers. We fixed the group size because the majority answer means different things for different numbers of readers, and the number of readings was not evenly distributed across methods: bilinear interpolation and no upscaling appeared in both experiments, so their positions were read a median of $21$ and $19$ times, against medians of $8$ to $11$ for the other methods. We set the group size at $5$ as the largest number of readers that retains almost all the data ($96.3\%$ of positions were read by at least five participants and the rest were excluded). For positions read by more than $5$ people, we averaged over every possible group of $5$ rather than picking one group at random. A position counted as read correctly if the majority answer (at least three votes out of five) was the true character, hallucinated if the majority was some other character, and as no consensus if no character held a majority.

We report each method's hallucination rate two ways: as a share of all its positions, and as a share of the positions that were misread. Confidence intervals were obtained by bootstrapping over plates. For each method we resampled its plates with replacement, $2\small{,}000$ times, recomputed both rates within each resampled set, and took the $2.5^{th}$ and $97.5^{th}$ percentiles of the resulting distributions.

\subsection*{Exploratory analysis: Counting versus identifying characters}

To further examine differences between upscaling methods, we broke performance down into two distinct parts: determining how many characters are present and determining which ones they are. We computed a \textit{miscount} rate (the percentage of attempted responses containing the wrong number of characters) and a \textit{misidentification} rate (among responses with the right number, the percentage with at least one error).

\section{Results}

\subsection*{Experiment 1: Conventional Upscaling}

\begin{figure}[t]
\centering
\includegraphics[width=\textwidth]{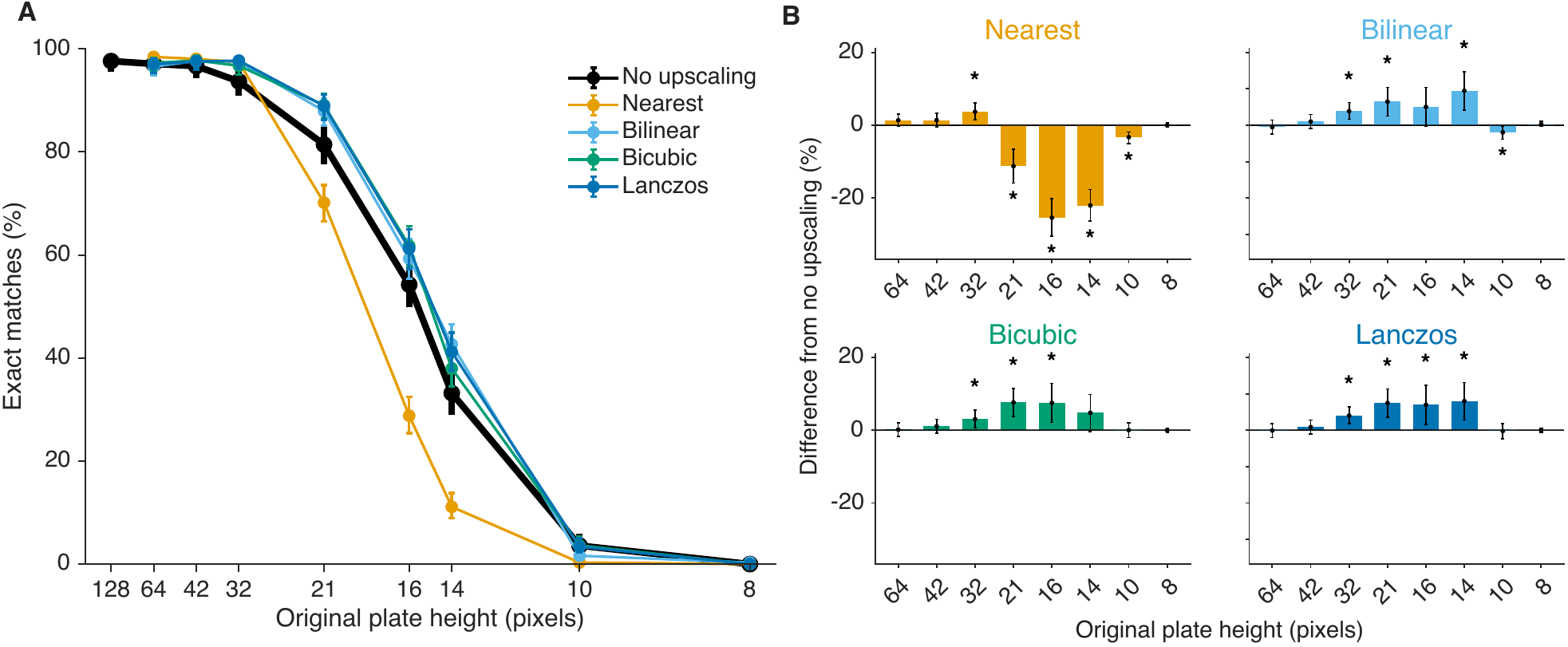}
\caption{Experiment 1: Percentage of exact matches by original plate height and upscaling method. \textbf{(A)} Percentage of trials on which the plate was transcribed exactly, plotted against the height of the plate before upscaling. Error bars are 95\% binomial intervals. \textbf{(B)} Difference from the no upscaling condition in percentage points at each height, with $95\%$ confidence intervals for the difference. The confidence intervals do not account for clustering by plate and are descriptive. Asterisks mark comparisons significant at $p_c<0.05$ after Benjamini--Hochberg correction, taken from the per-height comparisons in Table~\ref{tab:contrasts_conv_exact}. Only the heights from $32$ to $10$ pixels were tested; $64$, $42$ and $8$ pixels are plotted for completeness and carry no markers. Note that the no upscaling condition is the only one present at $128$ pixels, since no upscaling is applied at full resolution.}
\label{fig:conv_exact}
\end{figure}

We first examine how plate height affected performance when no upscaling was applied (Figure~\ref{fig:conv_exact}A, black line). Participants transcribed the plate exactly on $97.6\%$ of trials at full resolution ($128$ pixels tall). Performance was stable at $64$ and $42$ pixels tall, then fell off monotonically with no plates read correctly at $8$ pixels. Almost the entire transition from near-perfect to floor therefore occurred across a factor of three in plate height, between $32$ and $10$ pixels. Accordingly, the fitted model indicated that the odds of a correct reading in the no upscaling condition increased substantially with increasing plate height (Table~\ref{tab:glm_conv_exact}).

The effects of the upscaling methods on reading performance differed (Figure~\ref{fig:conv_exact}A, colored lines). While most methods were associated with improved performance at a range of heights, nearest neighbor interpolation was associated with overall reduced performance (Table~\ref{tab:glm_conv_exact}). In the fitted model, every method $\times$ height term was reliable, meaning that no method's benefit relative to no upscaling was constant across the height range. We therefore compared each method with no upscaling separately at each height (Figure~\ref{fig:conv_exact}B).

\begin{table}[t]
\centering
\caption{Experiment 1: mixed-effects logistic regression of correct response on upscaling method and $\log_2$ original plate height, with random intercepts for participant and plate. Random-effect standard deviations: participant 0.38, plate 1.58.}
\label{tab:glm_conv_exact}
{\footnotesize
\begin{tabular}{lrrrr}
\toprule
Term & OR & 95\% CI & $z$ & $p$ \\
\midrule
Intercept & 1.53 & [1.22, 1.92] & 3.70 & $<$0.001 \\
$\log_2$ height & 138.71 & [101.36, 189.83] & 30.81 & $<$0.001 \\
Bicubic & 1.91 & [1.61, 2.27] & 7.39 & $<$0.001 \\
Bilinear & 1.88 & [1.57, 2.24] & 7.00 & $<$0.001 \\
Lanczos & 2.00 & [1.68, 2.39] & 7.76 & $<$0.001 \\
Nearest & 0.31 & [0.26, 0.36] & -13.58 & $<$0.001 \\
Bicubic $\times$ $\log_2$ height & 2.60 & [1.62, 4.18] & 3.94 & $<$0.001 \\
Bilinear $\times$ $\log_2$ height & 3.59 & [2.18, 5.92] & 5.02 & $<$0.001 \\
Lanczos $\times$ $\log_2$ height & 3.04 & [1.87, 4.93] & 4.49 & $<$0.001 \\
Nearest $\times$ $\log_2$ height & 5.76 & [3.44, 9.66] & 6.66 & $<$0.001 \\
\bottomrule
\end{tabular}
}
\end{table}

The three smooth interpolators behaved almost interchangeably (Figure~\ref{fig:conv_exact}B, cool-colored bars). Bicubic, bilinear and Lanczos each modestly improved exact-match rates as compared to no upscaling for mid-range plate heights. Their advantage, in units of percent, narrowed for plates that were the shortest and tallest, as the no upscaling performance moved closer to the ceiling and floor. These results suggest that the benefit of conventional interpolation is real but limited, and confined to the range in which the plate is still marginally readable without it.

Nearest-neighbor upscaling, however, followed a different pattern (Figure~\ref{fig:conv_exact}B orange bars). Below $32$ pixels tall, it did not merely fail to help with reading but reduced legibility compared to no upscaling, by a margin up to $25.4\%$. Table~\ref{tab:contrasts_conv_exact} reports odds ratios for every method by height against no upscaling.

\begin{table}[t]
\centering
\caption{Experiment 1: each upscaling method against no upscaling at each original plate height, computed from the observed trial counts with standard errors clustered by plate. Odds ratios above one favor the method. $p_c$ is the $p$-value after Benjamini--Hochberg correction and * on an odds ratio marks $p_c<0.05$.}
\label{tab:contrasts_conv_exact}
{\footnotesize\setlength{\tabcolsep}{4pt}
\begin{tabular}{rrlr@{\hspace{10pt}}rlr}
\toprule
  & \multicolumn{3}{c}{Bicubic} & \multicolumn{3}{c}{Bilinear} \\
\cmidrule(lr){2-4}\cmidrule(lr){5-7}
Height (px) & OR & 95\% CI & $p_c$ & OR & 95\% CI & $p_c$ \\
\midrule
32 & 2.00* & [1.10, 3.63] & 0.034 & 2.69* & [1.37, 5.30] & 0.008 \\
21 & 1.86* & [1.25, 2.77] & 0.008 & 1.66* & [1.18, 2.34] & 0.008 \\
16 & 1.36* & [1.06, 1.76] & 0.025 & 1.23 & [0.96, 1.57] & 0.113 \\
14 & 1.23 & [0.96, 1.58] & 0.113 & 1.50* & [1.15, 1.96] & 0.008 \\
10 & 1.01 & [0.54, 1.89] & 0.971 & 0.44* & [0.21, 0.94] & 0.041 \\
\addlinespace
  & \multicolumn{3}{c}{Lanczos} & \multicolumn{3}{c}{Nearest} \\
\cmidrule(lr){2-4}\cmidrule(lr){5-7}
Height (px) & OR & 95\% CI & $p_c$ & OR & 95\% CI & $p_c$ \\
\midrule
32 & 2.89* & [1.41, 5.91] & 0.008 & 2.53* & [1.26, 5.10] & 0.015 \\
21 & 1.83* & [1.31, 2.56] & 0.002 & 0.54* & [0.39, 0.74] & 0.001 \\
16 & 1.33* & [1.03, 1.72] & 0.036 & 0.34* & [0.26, 0.44] & $<$0.001 \\
14 & 1.41* & [1.10, 1.80] & 0.011 & 0.25* & [0.17, 0.37] & $<$0.001 \\
10 & 0.94 & [0.47, 1.86] & 0.900 & 0.08* & [0.02, 0.32] & 0.002 \\
\bottomrule
\end{tabular}
}
\end{table}

Taken together, these findings suggest that conventional upscaling methods can improve legibility, but they do not always do so. The performance associated with nearest-neighbor interpolation, perhaps the most conservative of the four filters, illustrates that a transformation can be faithful to the data and still interfere with human legibility.

\subsection*{Experiment 2: Adaptive and AI-powered Upscaling}

\begin{figure}[t]
\centering
\includegraphics[width=\textwidth]{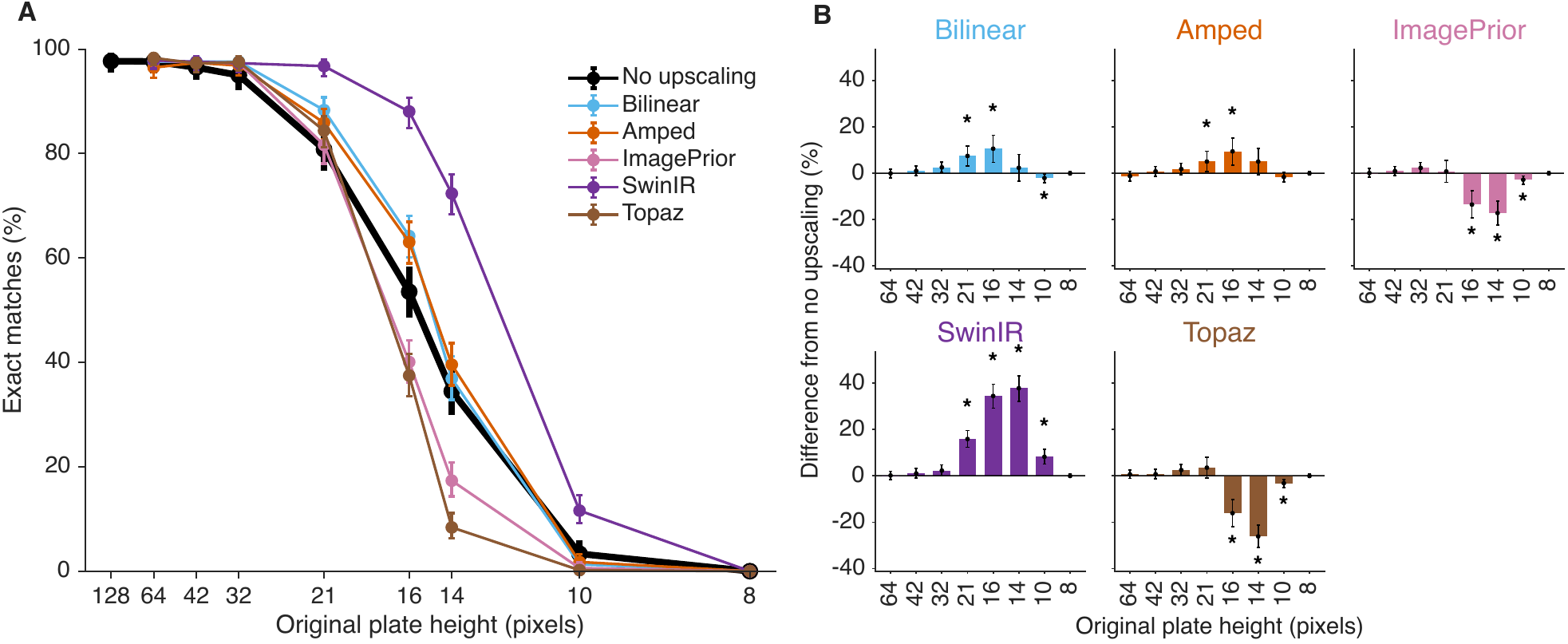}
\caption{Experiment 2: Percentage of exact matches by original plate height and upscaling method. \textbf{(A)} Percentage of trials on which the plate was transcribed exactly, plotted against the height of the plate before upscaling. Error bars are $95\%$ binomial intervals. \textbf{(B)} Difference from the no upscaling condition in percentage points at each height, with $95\%$ confidence intervals for the difference. The confidence intervals in do not account for clustering by plate and are descriptive. Asterisks mark comparisons significant at $p_c<0.05$ after Benjamini--Hochberg correction, taken from the per-height comparisons in Table~\ref{tab:contrasts_new_exact}. Only the heights from $32$ to $10$ pixels were tested; $64$, $42$ and $8$ pixels are plotted for completeness and carry no markers. Note that the no upscaling condition is the only one present at $128$ pixels, since no upscaling is applied at full resolution.}
\label{fig:new_exact}
\end{figure}

The results from the no upscaling condition in Experiment 1 replicated closely in this experiment (Figure~\ref{fig:new_exact}A, black line). The fitted model is given in~Table~\ref{tab:glm_new_exact}. As in Experiment 1 the odds of a correct reading rose steeply with height, and the method $\times$ height terms show that the effects of the upscaling methods differed over different plate heights (Figure~\ref{fig:new_exact}, colored lines and bars). We also included bilinear interpolation in this experiment in order to have a within-experiment comparison with a representative method from Experiment 1. Here again, the results closely matched with Experiment 1 (Figure~\ref{fig:new_exact}, blue line and bars). Note that the y-axis range in panel B differs between Figure~\ref{fig:conv_exact} and Figure~\ref{fig:new_exact}.

\begin{table}[t]
\centering
\caption{Experiment 2: mixed-effects logistic regression of correct response on upscaling method and $\log_2$ original plate height, with random intercepts for participant and plate. Random-effect standard deviations: participant 0.38, plate 1.60.}
\label{tab:glm_new_exact}
{\footnotesize
\begin{tabular}{lrrrr}
\toprule
Term & OR & 95\% CI & $z$ & $p$ \\
\midrule
Intercept & 1.54 & [1.22, 1.95] & 3.59 & $<$0.001 \\
$\log_2$ height & 148.97 & [105.12, 211.11] & 28.13 & $<$0.001 \\
Amped & 1.89 & [1.57, 2.28] & 6.61 & $<$0.001 \\
Bilinear & 1.80 & [1.49, 2.18] & 6.00 & $<$0.001 \\
ImagePrior & 0.63 & [0.53, 0.76] & -4.84 & $<$0.001 \\
SwinIR & 14.23 & [10.91, 18.55] & 19.61 & $<$0.001 \\
Topaz & 0.57 & [0.47, 0.68] & -5.83 & $<$0.001 \\
Amped $\times$ $\log_2$ height & 2.94 & [1.73, 5.01] & 3.98 & $<$0.001 \\
Bilinear $\times$ $\log_2$ height & 3.60 & [2.06, 6.29] & 4.50 & $<$0.001 \\
ImagePrior $\times$ $\log_2$ height & 5.31 & [2.99, 9.43] & 5.69 & $<$0.001 \\
SwinIR $\times$ $\log_2$ height & 5.36 & [3.01, 9.55] & 5.70 & $<$0.001 \\
Topaz $\times$ $\log_2$ height & 27.84 & [14.07, 55.12] & 9.55 & $<$0.001 \\
\bottomrule
\end{tabular}
}
\end{table}

Table~\ref{tab:contrasts_new_exact} reports odds ratios for every method by height against no upscaling. Amped Smart Resize, from the commercial forensic suite, produced gains comparable to the conventional interpolators of Experiment 1 (Figure~\ref{fig:new_exact}, red line and bars), while ImagePrior and Topaz reduced legibility. At $14$ pixels, for example, ImagePrior and Topaz reached exact match rates of only $17.3\%$ and $8.4\%$ against an exact match rate of $34.5\%$ for images with no upscaling. 

SwinIR was the only method tested in either experiment that substantially extended the range over which plates remained readable (Figure~\ref{fig:new_exact} purple line and bars). This method increased the exact match rate relative to no upscaling by as much as $37.9\%$. Its advantage persisted even at $10$ pixels. 

\begin{table}[t]
\centering
\caption{Experiment 2: each upscaling method against no upscaling at each original plate height, computed from the observed trial counts with standard errors clustered by plate. Odds ratios above one favor the method. $p_c$ is the $p$-value after Benjamini--Hochberg correction and * on an odds ratio marks $p_c<0.05$.}
\label{tab:contrasts_new_exact}
{\footnotesize\setlength{\tabcolsep}{4pt}
\begin{tabular}{rrlr@{\hspace{10pt}}rlr@{\hspace{10pt}}rlr}
\toprule
  & \multicolumn{3}{c}{Amped} & \multicolumn{3}{c}{Bilinear} & \multicolumn{3}{c}{ImagePrior} \\
\cmidrule(lr){2-4}\cmidrule(lr){5-7}\cmidrule(lr){8-10}
Height (px) & OR & 95\% CI & $p_c$ & OR & 95\% CI & $p_c$ & OR & 95\% CI & $p_c$ \\
\midrule
32 & 1.63 & [0.82, 3.21] & 0.184 & 2.14 & [1.05, 4.35] & 0.057 & 1.94 & [0.94, 4.02] & 0.104 \\
21 & 1.45* & [1.04, 2.02] & 0.049 & 1.80* & [1.29, 2.49] & 0.001 & 1.06 & [0.73, 1.53] & 0.772 \\
16 & 1.48* & [1.13, 1.94] & 0.009 & 1.56* & [1.19, 2.03] & 0.003 & 0.58* & [0.43, 0.78] & 0.001 \\
14 & 1.24 & [0.95, 1.63] & 0.137 & 1.11 & [0.83, 1.49] & 0.500 & 0.40* & [0.27, 0.58] & $<$0.001 \\
10 & 0.52 & [0.21, 1.29] & 0.184 & 0.38* & [0.17, 0.86] & 0.035 & 0.17* & [0.05, 0.61] & 0.013 \\
\addlinespace
  & \multicolumn{3}{c}{SwinIR} & \multicolumn{3}{c}{Topaz} & \multicolumn{3}{c}{} \\
\cmidrule(lr){2-4}\cmidrule(lr){5-7}
Height (px) & OR & 95\% CI & $p_c$ & OR & 95\% CI & $p_c$ &  &  &  \\
\midrule
32 & 1.91 & [0.90, 4.04] & 0.122 & 2.09 & [0.97, 4.48] & 0.086 &  &  &  \\
21 & 7.07* & [3.41, 14.67] & $<$0.001 & 1.28 & [0.88, 1.88] & 0.217 &  &  &  \\
16 & 6.39* & [4.09, 10.00] & $<$0.001 & 0.52* & [0.39, 0.70] & $<$0.001 &  &  &  \\
14 & 4.97* & [3.55, 6.97] & $<$0.001 & 0.17* & [0.11, 0.28] & $<$0.001 &  &  &  \\
10 & 3.81* & [1.99, 7.30] & $<$0.001 & 0.06* & [0.01, 0.32] & 0.003 &  &  &  \\
\bottomrule
\end{tabular}
}
\end{table}

\begin{figure}[t]
\centering
\includegraphics[width=0.7\textwidth]{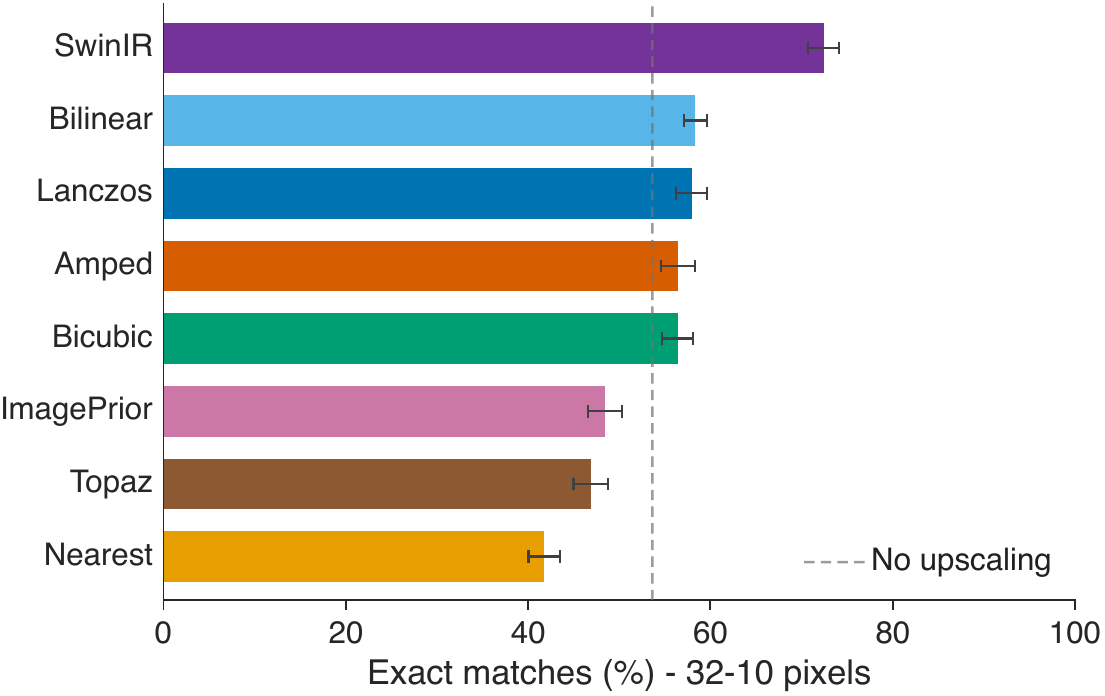}
\caption{Percentage of exact matches by upscaling method, pooled across both experiments and across the five analyzed plate heights ($32$, $21$, $16$, $14$ and $10$ pixels). Bars are trial-level proportions and error bars are $95\%$ binomial intervals on the pooled trials. The dashed line marks the no upscaling condition.}
\label{fig:all}
\end{figure}

The contrast between SwinIR on the one hand and ImagePrior and Topaz on the other hand is notable. All three synthesize high-frequency detail that is not present in the degraded image, albeit using different techniques. The results show that whether these inferred details help or harm legibility is likely architecture-specific. For a visual indication of the SwinIR performance compared to all other tested methods, Figure~\ref{fig:all} provides a descriptive summary of the performance of all methods across both experiments in terms of the average exact match rate pooled over plate heights from $32$ to $10$ pixels.

\subsection*{Hallucinated Characters}

\begin{figure}[t]
\centering
\includegraphics[width=\textwidth]{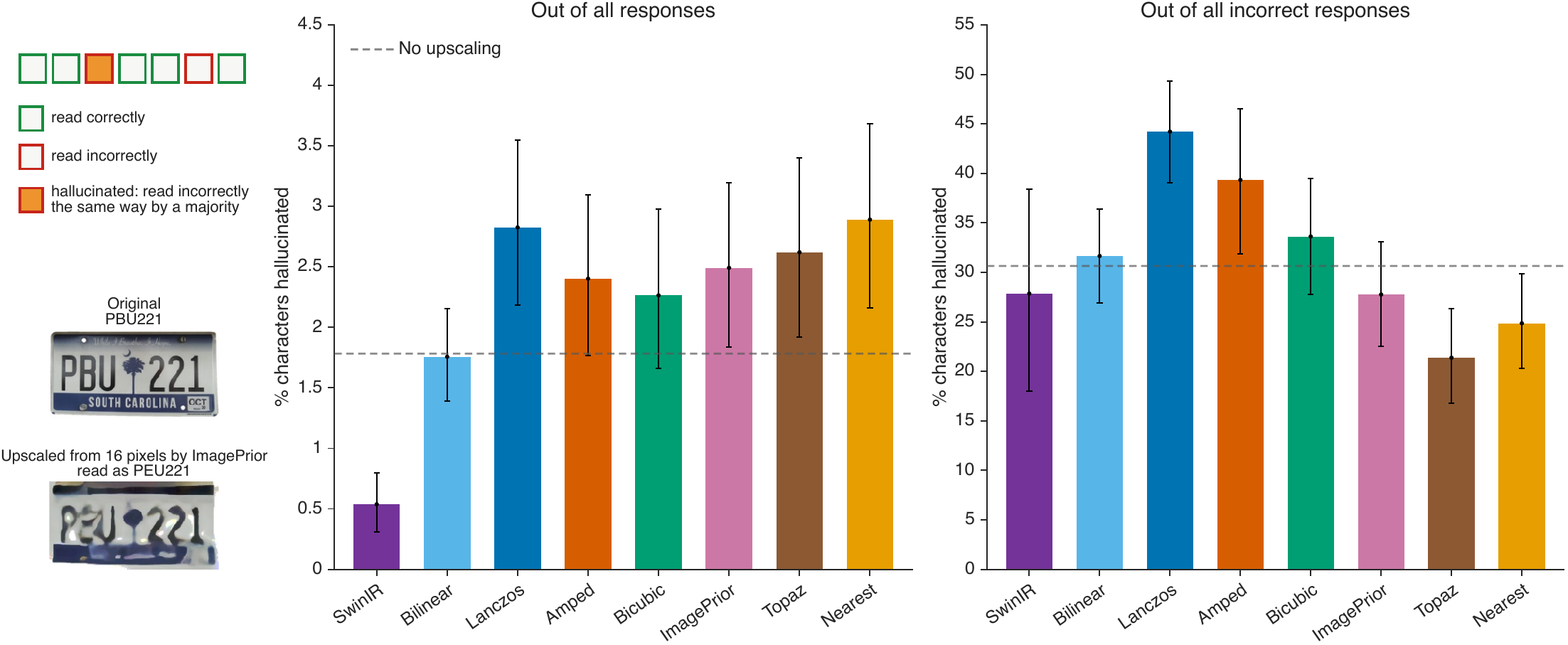}
\caption{Hallucinated character positions by upscaling method, pooled across both experiments at the five analyzed plate heights ($32$, $21$, $16$, $14$ and $10$ pixels). A position counts as hallucinated when a majority of five independent readers agreed on a character other than the true one. \textbf{Left panel:} Illustration of the classification, with green outlines marking positions read correctly, red outlines marking positions read incorrectly, and an orange fill marking the subset of those on which the readers agreed on the same wrong character. The bottom two images show an example of an unmodified plate image and one upscaled from 16 pixels tall with ImagePrior. \textbf{Middle panel:} Hallucinated positions as a percentage of all positions --- how much hallucination a method produces in absolute terms. \textbf{Right panel:} Hallucinated positions as a percentage of positions the readers failed to read correctly. Error bars are 95\% confidence intervals bootstrapped over plates; the dashed line marks the no-upscaling condition.}
\label{fig:hallucinations}
\end{figure}

One concern with upscaling techniques that rely on image statistics to add details is that the algorithm may synthesize a convincing character that was never present. This is a different failure from illegibility, and it is perhaps more concerning. An illegible plate is known to be uncertain while a hallucinated one may be mistaken for the truth. In an exploratory analysis, we therefore asked whether any of the upscaling methods resulted in substantially greater agreement among viewers' erroneous responses (Figure~\ref{fig:hallucinations}, left). An example of a common hallucination is shown as well: the South Carolina plate reading {\tt PBU 221} was consistently misread as {\tt PEU 221} when it was upscaled using ImagePrior.

These results suggest that hallucinations were not confined to the methods that synthesize detail (bars are ordered to match the best to worst performance ordering from Figure~\ref{fig:all}). In absolute terms (Figure~\ref{fig:hallucinations}, middle) $1.8\%$ of positions were hallucinated with no upscaling at all. Except for SwinIR ($0.5\%$), most methods were associated with a qualitatively similar hallucination rate. However, SwinIR also produced fewer errors overall, so next we examined hallucinations as a percentage of all errors (Figure~\ref{fig:hallucinations}, right). Of the positions that were not read correctly, $30.6\%$ were hallucinated under no upscaling. Qualitatively, the methods most likely to mislead when they failed were Lanczos and Amped, and those that were least likely to mislead were Topaz and nearest neighbor. But no methods resulted in hallucinations of less than $20\%$. 

Preliminarily, neither measure --- absolute or conditional hallucinations --- supports the hypothesis that AI-based upscaling is distinctively prone to hallucination. SwinIR, in particular, presents an interesting case: its conditional rate of $27.9\%$ is unremarkable, but it applies to so few failures ($1.9\%$ of positions) that its absolute rate is the lowest of those tested.

\subsection*{Counting Versus Identifying Characters}

\begin{figure}[t]
\centering
\includegraphics[width=\textwidth]{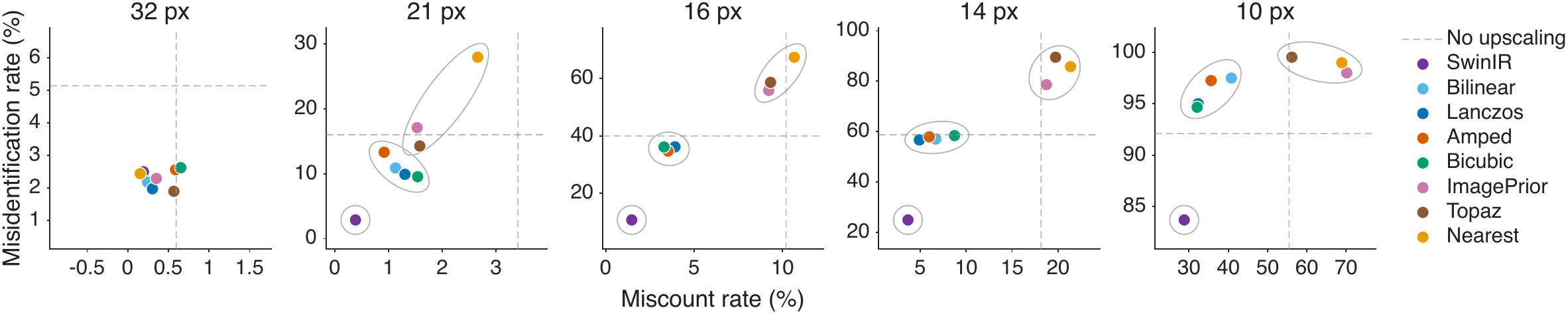}
\caption{Miscounting versus misidentifying. The horizontal axis is the percentage of attempted responses containing the wrong number of characters; the vertical axis is, among responses with the right number, the percentage with at least one error. The crosshair represents no-upscaling result. Ovals are illustrative, denoting groups of methods with different patterns.}
\label{fig:counting}
\end{figure}

Although hallucinations did not appear to differ strongly between the methods, it is clear from Figure~\ref{fig:teaser} that the different methods produce perceptually distinct outcomes. In a final analysis, we considered that a participant must work out how many characters are present (counting) before they can work out which ones they are (identifying). We asked whether these two aspects of performance may degrade differently. Shown in Figure~\ref{fig:counting} are the miscounts and misidentifications against each other for each method. The crosshair represents the no-upscaling result. A method to the left of the crosshair helped readers count characters, and a method below it helped them identify characters, and a method below and left of the crosshair helped on both.

For plate heights of $32$ pixels, every method resulted in miscounts on well under $1\%$ of trials and mis-identifications on only $2--5\%$. From $21$ to $10$ pixels tall, however, the methods separate into apparent clusters (see illustrative ovals). The three methods that tended to degrade performance in the main analysis (nearest neighbor, ImagePrior, and Topaz), appear to degrade identification but have relatively little effect on counting except at $10$ pixels. Relative to no upscaling, these methods did not produce more counting errors but did lead to more character misidentifications. The performance improvement associated with the smooth interpolators (bilinear, bicubic, Lanczos, and Amped) appears largely due to a reduction in miscounting, while identifications were only modestly affected. That is, much of the benefit of conventional interpolation may come from helping readers see how many characters there are. SwinIR is the only method that consistently improved both, suggesting that it is the only condition in our study that made individual characters substantially more identifiable rather than merely making the plate easier to parse.

\subsection*{Specialized Upscaling}

\begin{figure}[t]
\centering
\includegraphics[width=0.95\textwidth]{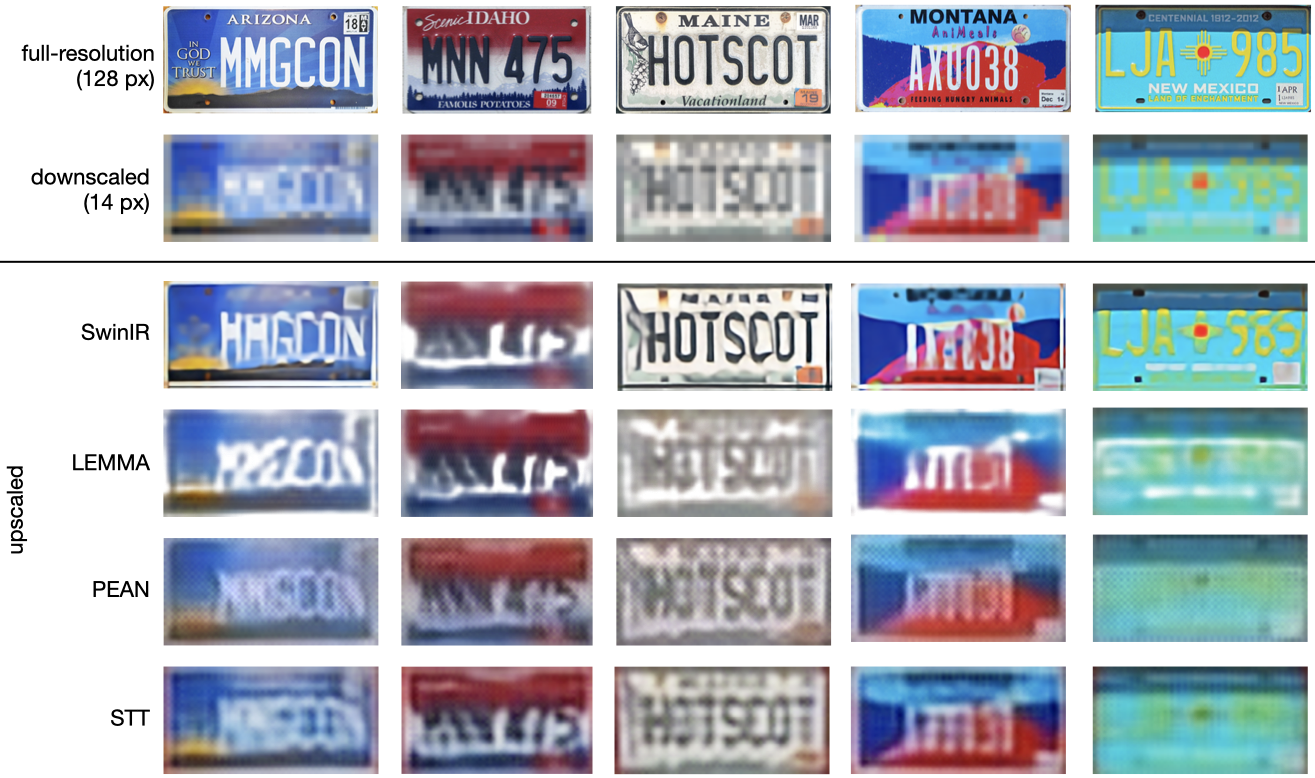}
\caption{Five representative license plates at their original full resolution (top row), these images downscaled to $14$ pixels tall (second row), and then upscaled to the original resolution using a general-purpose upscaling (SwinIR) and three Scene text image super resolution (STISR) techniques. As compared to SwinIR, these STISR techniques produce less legible results.}
\label{fig:STISR}
\end{figure}

In this study, we evaluated generic upscaling approaches applicable to any image, whether it contains text or not. Scene text image super resolution (STISR) techniques, on the other hand, are specifically designed to upscale images containing text. In a qualitative test, we evaluated three of these techniques with open-source trained models~\cite{Guo2023lemma, chen2021stt,zhao2024pean}. Because these models require an input of size $64 \times 16$ pixels ($4:1$ aspect ratio), we accommodated by padding our $2:1$ aspect ratio images with a mid-level gray border. Given the height restriction of $16$ pixels, we were only able to evaluate upscaling factors of $8$, $9$, $12$, and $16$. Shown in Figure~\ref{fig:STISR} are representative examples of the results compared to no upscaling (second row) and SwinIR (third row), the best performing general-purpose upscaling. We see that qualitatively, these STISR techniques perform worse in terms of human readability.

We hypothesize several factors contribute to this relatively poor performance. First, these methods were trained and evaluated against high-contrast images of printed text on a largely monotone background~\cite{wang2020textzoom}. These techniques, therefore, are likely challenged by the license plates in our dataset that feature more complex layouts and patterned backgrounds. Second, one of these techniques (LEMMA) uses textual context to infer character identity. This context is likely to create a bias to predicting dictionary words which is not applicable to most license plates. Third, and lastly, with a forced input size of $64 \times 16$ pixels, these approaches are not applicable to a broad range of image resolutions. Given these limitations, and the qualitative results observed in Figure~\ref{fig:STISR}, we did not pursue these techniques further.

\section{Discussion}

Across eight upscaling methods, we found that the empirical ability to help a human reader decipher a license plate varied substantially, but did not strictly follow from how sophisticated, modern, or explainable they were.

Surprisingly, several methods were worse than doing nothing. In particular, nearest-neighbor interpolation preserves every original pixel exactly and invents nothing, yet it made many of the legible plates illegible. The fact that an interpolation method adds no new information is therefore no guarantee that it does no harm: what reaches the reader is not the information content of the image but its appearance. A visible sampling grid, for example, may interfere with character identification as effectively as a fabricated stroke does. Two learned upscaling methods --- ImagePrior and Topaz --- produced qualitatively less dramatic degradation in legibility (Figure~\ref{fig:all}).

Of the upscaling methods that helped, almost all but one helped only modestly. Bilinear, bicubic, Lanczos and a commercial forensic tool (Amped Smart Resize) each supported about $5--10$ percentage point increases in the middle of the plate size range. The counting analysis suggests that they did so mainly by helping readers determine how many characters were present rather than by making the characters themselves more discriminable. 

SwinIR is the exception: it raised exact-match rates by up to $38$ points, extended legibility a full step down the downscaling ladder, and did so by improving character identification. Why SwinIR should be so much better than the other learned methods is not something the current experiments can settle. Several differences are candidates. None of the techniques tested here are designed specifically for text or license plates. It is possible that the specific statistics of license plate imagery happen to suit some architectures and training data better than others. Thus, when general-purpose upscalers are being considered for high-stakes use cases, direct testing of human performance may be essential. 

There is a smaller body of work that seeks to directly infer character identity without explicitly upscaling the image~\cite{lorch2019forensic,lorch2022compression}. While these approaches are worthy of examination in a general forensic setting, our focus here is on human readability, and hence these techniques were not part of our evaluation.

In forensic and legal settings, the question of whether AI-powered enhancement of visual media is helpful is gaining increasing importance. Standard image-processing algorithms for enhancement have long been a mainstay of forensic practice~\cite{swgde2022resizing}. As such, the act of ``making up pixels'' does not inherently violate legal precedent. But with new techniques for image enhancement becoming available, the courts will have to decide what is acceptable. In 2024, for example, the defense team in \textit{State of Washington v.\ Joshua Puloka} attempted to admit video evidence that had been ``enhanced by artificial intelligence'' in support of a claim of self-defense in a murder trial~\cite{puloka2024}. The original video was of poor quality, and the enhanced version had been processed using a commercial video editing tool with AI-powered technology (Topaz Labs AI). The court excluded it, finding that the method had not achieved general acceptance in the scientific community as required under \textit{Frye}~\cite{frye1923}. The original unenhanced video was admitted, and Puloka was found guilty. It is unlikely, however, that this is the last time AI-enhanced imagery will make its way into the courtroom. The federal courts and most states follow the Daubert standard for admissibility~\cite{daubert1993}, under which a trial judge also assesses reliability directly, including determining whether the method has a known error rate. For an enhancement offered to help a juror or a forensic examiner read something, the error rate that matters is the rate at which human viewers read the output correctly. These results aim to provide data to support these types of legal judgments.

Several limitations are important to consider. Amongst the range of visual imagery that may be upscaled or otherwise enhanced, license plates are just one case, selected partially due to their ease of measurement: the ground truth is discrete, unambiguous and known, and accuracy can be scored character by character. Faces, gait, clothing and actions --- content often also at issue in surveillance footage --- may not offer such ground truth, and the effects of upscaling on them may differ. In addition, our stimuli were clean, evenly lit, frontal crops subjected to a single controlled degradation, whereas real evidentiary imagery may simultaneously contain motion blur, compression, low light and oblique viewing angles. Additionally, we tested static images, not video, and video affords a viewer and an enhancement tool multiple views of the same object that can facilitate reading. Finally, participants read plates in a browser at their own display settings (which we verified but did not control). They only saw a fixed resolution and their viewing distance was unknown. A forensic examiner working with an interactively zoomable image at length may perform differently.

The natural extensions follow from those limits. The experimental paradigm used here to study legibility may be adapted to study perception of faces and actions in video, where the ground truth is harder, but not impossible, to establish. It could also be applied to compounded degradations. With more research in this domain, we can move towards understanding what distinguishes an enhancer that helps from one that harms. Until that question has an answer, the only safe assumption is that an enhancement method helps a human reader exactly as much as it has been shown to.

\begin{acks}
The authors thank Christine Corry and Aakarsh Gopisetty for help with sourcing code and image generation. Claude was utilized to draft sections of this Work, including text, tables, graphs, and citations.
\end{acks}

\bibliographystyle{ACM-Reference-Format}
\bibliography{references}

\end{document}